\pdfoutput=1

\documentclass[11pt]{article}

\usepackage[]{ACL2023}

\usepackage{times}
\usepackage{latexsym}

\usepackage[T1]{fontenc}

\usepackage[utf8]{inputenc}

\usepackage{microtype}

\usepackage{inconsolata}

\usepackage{hyperref}
\usepackage{url}

\usepackage{graphicx}
\usepackage{colortbl}
\usepackage{bbm}
\usepackage{amsfonts}
\usepackage[normalem]{ulem}
\useunder{\uline}{\ul}{}
\usepackage{amssymb}
\usepackage{amsfonts}
\usepackage{framed}
\usepackage{color}
\usepackage{multicol}
\usepackage{multirow}
\usepackage{makecell}
\usepackage{amsmath}
\usepackage{booktabs}
\usepackage{array}
\usepackage{paralist}
\usepackage{algorithm}
\usepackage{bbm}
\usepackage{dsfont}
\usepackage{adjustbox}
\usepackage{diagbox}
\usepackage{algpseudocode}
\usepackage{multicol}
\usepackage{wrapfig}
\usepackage{enumitem}
\usepackage{caption}
\usepackage{fontawesome}

\AtBeginDocument{\setlength\abovedisplayskip{4pt}}
\AtBeginDocument{\setlength\belowdisplayskip{4pt}}

\title{Multi-Source Test-Time Adaptation as Dueling Bandits \\ for Extractive Question Answering}

\author{Hai Ye \ \ \ Qizhe Xie \ \ \ Hwee Tou Ng \\ 
Department of Computer Science, National University of Singapore \\
\texttt{\{yehai,qizhex,nght\}}\texttt{@comp.nus.edu.sg}
}

\begin{document}
\maketitle
\begin{abstract}

In this work, we study multi-source test-time model adaptation from user feedback, where $K$ distinct models are established for adaptation. To allow efficient adaptation, we cast the problem as a stochastic decision-making process, aiming to determine the best adapted model after adaptation. We discuss two frameworks: multi-armed bandit learning and multi-armed dueling bandits. Compared to multi-armed bandit learning, the dueling framework allows pairwise collaboration among $K$ models, which is solved by a novel method named Co-UCB proposed in this work. Experiments on six datasets of extractive question answering (QA) show that the dueling framework using Co-UCB is more effective than other strong baselines for our studied problem\footnote{Code of the paper is available at \url{https://github.com/oceanypt/Multi-source-TTA}.}.  


\end{abstract}

\section{Introduction}

Large language models~(LLMs) can be fine-tuned or prompted with texts to achieve good performance in NLP tasks~\cite{DBLP:conf/naacl/DevlinCLT19,brown2020language,ouyang2022training}. However, because of the unexpected distribution shift at test time, the effectiveness of LLMs can degenerate ~\cite{wang2021measure}. They may also generate outputs that are untrustworthy or toxic and fail to meet user expectations~\cite{ouyang2022training}. 
One critical issue that we need to address is to improve the generalization ability of LLMs. Recent research on test-time adaptation~(TTA) suggests a possible way to do this, by continually updating the deployed model with target data from an arbitrary test distribution~\cite{wang2020tent}.

\begin{figure}[t]
\setlength{\abovecaptionskip}{0.2cm}
\setlength{\belowcaptionskip}{-0.3cm}
    \centering
    \includegraphics[width=7cm]{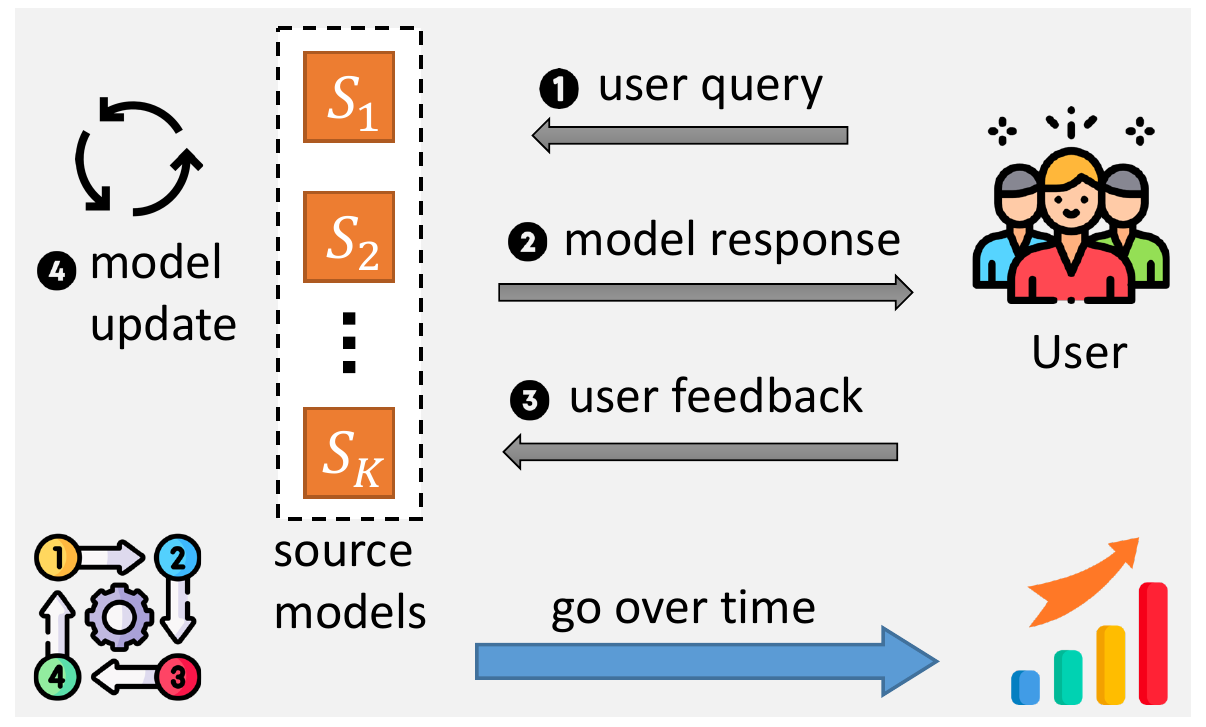}
    \caption{The illustration of multi-source test-time adaptation from user feedback studied in this work. Each model is trained from a distinct source domain. With unlabeled test data, models are adapted online from user feedback.}
    \label{fig:ill-multi-source-tta}
\end{figure}


Interacting with users is important during test-time adaptation. First, user feedback allows the model to better align with humans~\cite{stiennon2020learning,ouyang2022training}. 
Users can directly teach the model to learn by interaction so that the model can be better trained to follow human instructions and reduce the generation of toxic and harmful content. 
Besides, obtaining feedback from users can also reduce the cost of data annotation by experts, and the collected data will be more in line with the distribution of the users~\cite{DBLP:conf/emnlp/NguyenDB17,gao2022simulating}, which makes the adaptation more economical and effective. 

Leveraging multiple learned models of tasks is also important for TTA. 
As in previous work, utilizing multiple known tasks helps the model better learn new tasks (or distributions), such as meta-learning~\cite{hospedales2021meta} 
and multi-source domain adaptation~\cite{ramponi2020neural}. 
To take advantage of known tasks, compared to reusing task data, directly using their learned models has gained popularity recently~\cite{pfeiffer2020adapterfusion,wang2021efficient}, which is much cheaper for online adaptation and has better data privacy protection~\cite{kundu2020universal}. 
Recent work on lightweight tuning empowers LLMs to store knowledge of a large number of tasks cheaply~\cite{DBLP:conf/icml/HoulsbyGJMLGAG19,liu2021pre}. 
Platforms like Huggingface~\cite{DBLP:journals/corr/abs-1910-03771} also allow users to share locally trained models, promoting a large amount of knowledge stored as models in the cloud. So, it has become more critical for TTA to adapt from multiple learned models of tasks. 



Based on the above discussion, we propose to study an important but under-explored problem --  multi-source test-time adaptation from user feedback -- where $K$ source models are given, each trained from a distinct source domain, to adapt to a new target domain~(Figure~\ref{fig:ill-multi-source-tta}). 
Previous work on leveraging multiple knowledge sources is to learn an ensemble~\cite{DBLP:conf/emnlp/GuoSB18,DBLP:conf/cvpr/AhmedRPOR21}, which means jointly accessing all the models is needed for training. Due to its high cost, it is not suitable for real-time updates required by TTA. In order to adapt efficiently, we turn this problem into a stochastic decision-making process that trades off model exploration and exploitation. We aim to determine the best adapted model that can perform well in the target domain. 

We formulate the problem in two frameworks: multi-armed bandit learning and multi-armed dueling bandits~\cite{kuleshov2014algorithms}. 
Bandit learning samples one source model each time to receive binary feedback~(\faThumbsOUp~or~\faThumbsODown)~($\S$\ref{sec:UCB}). 
However, it lacks collaboration among sources and can result in a sub-optimal adapted model. In order not to introduce too much cost, pairwise collaboration between models is explored in dueling bandits~\cite{DBLP:conf/colt/YueBKJ09}, where two distinct source models are chosen each time for dueling with user preference feedback~(e.g., \faChevronCircleRight/\faChevronCircleLeft)~($\S$\ref{sec:Co-UCB}). A novel method, Co-UCB, is proposed to allow collaborative updates. 

We choose to study the task of extractive question answering (QA), since there are large datasets in different domains that can be used~\cite{fisch2019mrqa}. More importantly, extractive QA is suitable for eliciting users to leave feedback, 
since the surrounding context around predicted answer spans can help users to verify the answers. \citet{gao2022simulating} has simulated user feedback for TTA in extractive QA, but not in the multi-source scenario. 
Following previous work~\cite{gao2022simulating}, we simulate user feedback with the annotated answer spans. We conduct our simulation experiments on the MRQA benchmark~\cite{fisch2019mrqa}, where six domains of extractive QA are studied. We compare the two proposed frameworks to assess their effectiveness and reveal the differences. We also look into the effect of noisy preference feedback. 

Our contributions in this work are as follows:
\begin{compactitem}
    \item We are the first to study multi-source test-time adaptation from user feedback; 
    \item We propose a novel formulation of the problem as dueling bandits and solve it by a new method;
    \item Preference feedback is discussed for extractive QA for the first time; and
    \item Extensive experiments and analysis are conducted to verify our method. 
\end{compactitem}

\begin{figure*}[t]
\setlength{\belowcaptionskip}{-0.1cm}
    \centering
    \includegraphics[width=\textwidth]{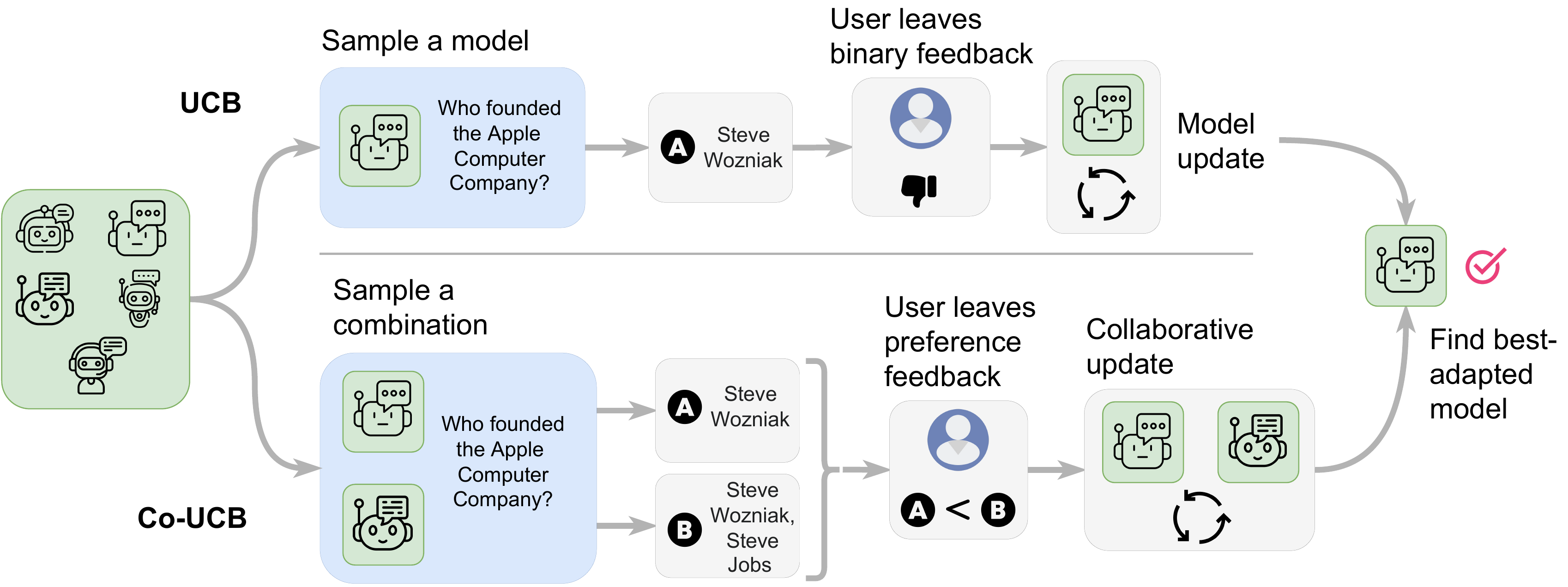}
    \caption{The illustration of UCB and Co-UCB for multi-armed bandit learning and dueling bandits respectively. For UCB, one model is sampled and receives binary feedback. For Co-UCB, two models are chosen each time and preference feedback is made by the user, then the models are updated.}
    \label{fig:co-ucb}
\end{figure*}

\section{Related Work}
\noindent{\textbf{Domain Adaptation.}} \ Adapting from source domain(s) to target domain(s) is important for generalized machine learning~\cite{ramponi2020neural}. Test-time adaptation~(TTA) attracts much attention recently, which adapts with the test data on the fly~\cite{pmlr-v119-sun20b,DBLP:conf/nips/IwasawaM21,wang2020tent,DBLP:journals/corr/abs-2302-04618}. TTA is more suitable for domain generalization since it needs no source data, which is different from unsupervised domain adaptation~(UDA)~\cite{ramponi2020neural,ye2020feature}. Multi-source DA is a more challenging problem than UDA, since it needs to determine suitable source knowledge for adaptation~\cite{DBLP:conf/emnlp/GuoSB18,DBLP:conf/aaai/GuoPB20}. Multi-source TTA has not been explored in NLP. Different from multi-source DA, multi-source TTA has no access to the source training data, and only the source models are given, which makes it more challenging to exploit useful sources for adaptation.

\noindent{\textbf{Learning from Human Feedback.}} \ Human feedback is a useful signal to refine the model outputs and adapt to new domains~\cite{gao2022simulating}, follow human instructions~\cite{ouyang2022training}, etc. Human feedback has been explored for different NLP tasks such as machine translation~\cite{DBLP:conf/emnlp/NguyenDB17,DBLP:conf/acl/KreutzerR19,DBLP:conf/acl/MendoncaRCSS20}, semantic parsing~\cite{DBLP:conf/acl/RiezlerL18,yao2020imitation,elgohary2021nl}, document summarization~\cite{DBLP:conf/emnlp/GaoMG18,stiennon2020learning}, question answering~\cite{DBLP:conf/emnlp/KratzwaldFS20,gao2022simulating}, and dialogue systems~\cite{DBLP:journals/corr/abs-2208-03188}. In particular, learning from human feedback has gained a lot of interests recently in the context of alignment of large language models (LLMs)~\cite{stiennon2020learning, ouyang2022training, OpenAI2023GPT4TR}. Fundamentally, alignment research is necessary and appealing from two aspects: (1) Alignment enables a model to go beyond supervised learning~\citep{stiennon2020learning} (2) Alignment leads to a safer system~\citep{OpenAI2023GPT4TR}. The proposed co-UCB could potentially be used for alignment in future work.

\section{Preliminaries}
\subsection{Problem Definition}
\noindent{\textbf{Multi-source TTA.}} \ We study multi-source test-time domain adaptation from $K$ source models by interacting with users, where each model is trained from a distinct source domain. Test-time data is from a target domain $X$ which is unlabeled. We focus on online adaptation where the test data $\mathbf{x} \sim X$ comes as a stream and the model is continually updated with newly emerged test data at each time $t$\footnote{Note that our method can be easily applied in the offline scenario.}. The parameters of each model at time $t$ are inherited from the last time $t-1$. 
At each time $t$, we obtain the test data $\mathbf{x}_t \sim X$ which is the concatenation of the question $\mathbf{q}_t$ and the passage $\mathbf{d}_t$. The prediction ${y}_t$ is a pair of start and end positions over the passage denoted as $\langle y_t^{(1)}, y_t^{(2)} \rangle$. Following previous work~\cite{rajpurkar2016squad}, we use cross-entropy loss to update the model which is:
\begin{equation}
    \mathcal{L}_t = -\Big(\log(p_t^{(1)}) + \log(p_t^{(2)}) \Big) / 2 
    \label{eq:cross-entropy-loss}
\end{equation}
where $p_t^{(1)}$ and $p_t^{(2)}$ are the probabilities of the predicted start $y_t^{(1)}$ and end $y_t^{(2)}$ positions respectively. 

\noindent{\textbf{Motivation.}} It is costly to learn an ensemble of $K$ sources, since it has at least $K$ times the training and inference costs, and even $K$ times the parameters of a single source model~\cite{DBLP:conf/emnlp/GuoSB18,DBLP:conf/cvpr/AhmedRPOR21}. In order to adapt efficiently, we cast the problem as a stochastic decision-making process, where we aim to determine the best adapted model that can perform well in the target domain through user interaction. 

\noindent{\textbf{Frameworks.}} \ We first formulate the problem as multi-armed bandit learning~\cite{kuleshov2014algorithms} and show how to solve it with Upper Confidence Bound~(UCB)~\cite{agrawal1995sample,DBLP:journals/ml/AuerCF02}~($\S$\ref{sec:UCB}). We further discuss multi-armed dueling bandits to address the drawback of bandit learning, and propose a novel method Co-UCB~($\S$\ref{sec:Co-UCB}).

\subsection{Background}

\noindent{\textbf{Multi-Armed Bandit Learning.}} \ The learning of multi-armed bandits~(MAB) is a stochastic and iterative problem~\cite{DBLP:conf/ijcai/SuiZHY18}, which repeatedly selects \emph{a model} from $K$ sources. Each selected model receives a reward from the user. After $T$ iterations, the goal of MAB is to minimize the cumulative regret compared to the best model:
\begin{equation}
    \mathcal{R}^{\text{MAB}}(T) =  \sum_{t=1}^T \Big [ \mu^* - \mu(a_t) \Big ]
    \label{eq:mab-definition}
\end{equation}
where $a_t$ is the action at time $t$ and $\mu(a)$ is the expected reward of the action $a$. $\mu^*$ is the expected reward of the best model. 

\noindent{\textbf{Multi-Armed Dueling Bandits.}} \ In the multi-armed dueling bandits~(MADB) problem, \emph{two distinct models} are sampled among the $K$ models~\cite{DBLP:conf/colt/YueBKJ09}. Also, the user needs to indicate a preference over the two selected models. In each comparison, a model $a_i$ is preferred over $a_j$ with the probability $P(a_i > a_j)$, which is equal to $\epsilon(a_i, a_j) + 1/2$ where $\epsilon(a_i, a_j) \in (-1/2, 1/2)$. Suppose two models $a^{(i)}_t$ and $a^{(j)}_t$ are sampled at time $t$, and $a^*$ is the overall best model. We define the cumulative regret at time $T$ as:
\begin{equation}
    \mathcal{R}^{\text{MADB}}(T) = \sum_{t=1}^{T} \Big [ \epsilon (a^*, a^{(i)}_t ) + \epsilon ( a^*, a^{(j)}_t )  \Big ] 
\end{equation}
which is a strong version discussed in \citet{DBLP:conf/colt/YueBKJ09}. It is the proportion of users who prefer the best model over the selected ones each time. 


\begin{algorithm}[t!]
  \begin{algorithmic}[1]
  \Require $K$ source models.
  \State $\bar{{\boldsymbol{\mu}}} \gets \mathbf{0}^K$, $\mathbf{n} \gets \mathbf{0}^K$, $N \gets 0$;
  \For{ $\mathcal{B}_t \in X$    }
    \State $k \gets {\arg\max}_j \ \bar{\mu}_j + \sqrt{2\ln(N) / { {n}_j }} $;
    \State Obtain the reward $\mathbf{r}_t$ for model $k$; 
    \State $\bar{\mu}_k \gets (\bar{\mu}_k {n}_k + \mathbf{r}_t^{\top}\mathbf{r}_t) / ( {n}_k + |\mathcal{B}_t| ) $; 
    \State ${n}_k \gets {n}_k + |\mathcal{B}_t|$; $N \gets N + |\mathcal{B}_t|$;
    \State \parbox[t]{\dimexpr\linewidth-\algorithmicindent}{Update model $k$ with loss  $\mathbf{r}_t^{\top} \mathcal{L}_t / |\mathcal{B}_t|$; \textcolor{gray}{// $\mathcal{L}_t$ is as Eq.~\ref{eq:cross-entropy-loss} shows.}}
    
  \EndFor
  \State \textbf{Return:} $k \gets {\arg\max}_j \ \bar{\mu}_j + \sqrt{2\ln(N) / { {n}_j }} $.
  \end{algorithmic}
  \caption{UCB for $K$-armed bandit learning}
  \label{algo:ucb}
\end{algorithm}

\section{UCB for Bandit Learning}\label{sec:UCB}
As illustrated by Figure~\ref{fig:co-ucb}, we apply UCB for multi-armed bandit learning, whose pseudo-code is shown in Algorithm~\ref{algo:ucb}.

\noindent{\textbf{Action.}} \ At each time $t$, the source model $k$ is selected from $K$ source models which maximizes $\bar{\mu}_k + \sqrt{\frac{2\ln(N)}{n_k}}$, where $\bar{\mu}_k$ represents the average reward obtained for the model $k$ by attempting it for $n_k$ times, and $N$ is the number of all test data instances received so far. $\sqrt{\frac{2\ln(N)}{n_k}}$ represents the confidence interval to the action $k$, and a larger one means more uncertainty about the action, intending to explore the action more. As training proceeds, the policy becomes more confident about each action.

\noindent{\textbf{Simulated Binary Feedback~(\faThumbsOUp~or~\faThumbsODown).}} \ For each input, the selected model will first predict its answer, then the user leaves the feedback to the prediction. Here, we use binary feedback since it is simple enough for the user to provide and has often been used in previous work~\cite{DBLP:conf/emnlp/KratzwaldFS20,gao2022simulating}. At each time $t$, a batch input $\mathcal{B}_t \sim X$ is obtained for training, which is passed to the model $k$ to obtain the predictions. 

\noindent{\textbf{Reward.}} \ With a batch of predicted answers, the model $k$ will obtain a vector of simulated reward $\mathbf{r}_t \in \{0, 1\}^{|\mathcal{B}_t|}$ decided by the user. For each data instance in the batch, we follow \citet{gao2022simulating} to calculate the simulated reward by comparing the predicted answer to the annotated span, where an index-wise exact match is used. If both the predicted start and end positions exactly match the annotated positions, the reward is $1$; otherwise, $0$.

\noindent{\textbf{Model Update.}} \ After obtaining the reward, the model $k$ will be updated with a reward-enhanced loss, where the task-specific cross-entropy loss $\mathcal{L}_t$~(in Eq.~\ref{eq:cross-entropy-loss}) will be multiplied by the reward $\mathbf{r}_t$. 

\noindent{\textbf{Inference.}} After enough iterations, the best adapted model can be found to perform well in the target domain as line $9$ of Algorithm~\ref{algo:ucb} shows.

\section{Collaborative UCB for Dueling Bandits}\label{sec:Co-UCB}
\subsection{Co-UCB}
\noindent{\textbf{Motivation.}} \ Since only one model is accessed each time in bandit learning, unlike ensemble learning~\cite{DBLP:conf/emnlp/GuoSB18}, it cannot make use of the collaboration among sources during adaptation. To address such a drawback and not incur much extra training cost, we exploit the pairwise collaboration among $K$ sources, where each time two distinct models will be sampled for joint learning. After adaptation, we also keep the best model for inference, to have the same cost as bandit learning. 

Sampling pairs of models can be formulated as multi-armed dueling bandits~(MADB) as discussed above. However, previous work on MADB only aims to determine the best model~\cite{DBLP:conf/colt/YueBKJ09,DBLP:conf/icml/ZoghiWMR14,DBLP:conf/uai/SuiZBY17}, so we further propose a novel method which is Collaborative UCB~(Co-UCB) to let a pair of models collaborate, whose pseudo-code is presented in Algorithm~\ref{algo:co-ucb}, and illustrated by Figure~\ref{fig:co-ucb}.


\noindent{\textbf{Action.}} \ At each time $t$, with $K$ source models, we construct $C_2^{K}$ combinations for selection, where each combination is denoted by a pair of model indices $\langle i, j \rangle$ ($i < j$). The combination $\langle i, j \rangle$ selected at time $t$ should maximize $ (\bar{\mu}_{i} + \bar{\mu}_j)/2 + \sqrt{2\ln{(N)} / n_{i,j}}$, where $\bar{\mu}_i$ and $\bar{\mu}_j$ are the average reward obtained up to time $t$ of model $i$ and $j$ respectively, and $n_{i,j}$ is the number of combinations $\langle i, j \rangle$ explored so far. $N$ is the total number of test data instances received until now. 

Take model $i$ for example. The reward of exploiting model $i$ represents how well model $i$ can beat the other models during dueling. The average reward $\bar{\mu}_i$ is calculated as follows:
\begin{equation}
    \bar{\mu}_i = \sum_{k=1, k \ne i}^{K} r_{i,k} / \sum_{k=1, k \ne i}^K n_{i, k}
\end{equation}
where $r_{i,k}$ denotes the overall reward that the model $i$ received by dueling with model $k$ and $n_{i,k}$ denotes the number of times model $i$ duels with model $k$. 

In each selection, to calculate the average reward $(\bar{\mu}_i + \bar{\mu}_j)/2$ for the combination $\langle i, j \rangle$, we expect $\langle i, j \rangle$ to be the most worthy action~(exploration-and-exploitation trade-off), where $i$ and $j$ can mostly beat the rest of the models, which means they are the two strongest models among the $K$ sources so that they can better collaborate to improve them. 

\begin{algorithm}[t!]
  \begin{algorithmic}[1]
  \Require $K$ source models.
  \State ${\bar{\boldsymbol{\mu}}} \gets \mathbf{0}^K$, $\mathbf{n} \gets \mathbf{0}^{K \times K}$, $N \gets 0$;
  \For{ $\mathcal{B}_t \in X$    }
    \State \text{\small{$ \langle i, j \rangle \gets \underset{i,j; i< j}{\arg\max} \ (\bar{\mu}_i + \bar{\mu}_j)/2 + \sqrt{2\ln(N) / { {n}_{i,j} }}$;}} 
    \State \parbox[t]{\dimexpr\linewidth-\algorithmicindent}{Obtain the rewards $\mathbf{r}_i$ and $\mathbf{r}_j$ for model $i$ and $j$ respectively as in Eq.~\ref{eq:preference-reward};} 
    \State \parbox[t]{\dimexpr\linewidth-\algorithmicindent}{$\bar{\mu}_i \gets (\bar{\mu}_i \sum_{k} {n}_{i,k} + \mathbf{r}^{\top}_i\mathbf{r}_i) / ( \sum_{k}{n}_{i,k} + |\mathcal{B}_t| ) $;} 
    \State ${n}_{i,j} \gets {n}_{i,j} + |\mathcal{B}_t|$; 
    \State \parbox[t]{\dimexpr\linewidth-\algorithmicindent}{$\bar{\mu}_j \gets (\bar{\mu}_j \sum_{k} {n}_{j,k} + \mathbf{r}^{\top}_j\mathbf{r}_j) / ( \sum_{k}{n}_{j,k} + |\mathcal{B}_t| ) $; }
    \State ${n}_{j,i} \gets {n}_{j,i} + |\mathcal{B}_t|$; 
    \State $N \gets |\mathcal{B}_t|$;
    \State \parbox[t]{\dimexpr\linewidth-\algorithmicindent}{Update the models $i,j$ by loss $\mathcal{L}_t$ as in Eq.~\ref{eq:collaborative-update};}
  \EndFor
  \State \textbf{Return:} \small{$k \gets \underset{j}{\arg\max} \ \bar{\mu}_j + \sqrt{2\ln(2N) /  \sum_{k}{ {n}_{j,k} }} $}.
  \end{algorithmic}
  \caption{Co-UCB for $K$-armed dueling bandits}
  \label{algo:co-ucb}
\end{algorithm}

\noindent{\textbf{Simulated Preference Feedback.~(e.g., \faChevronCircleRight/\faChevronCircleLeft)}} \ Since for each input, the user will receive two answer candidates instead of one, the binary feedback used in bandit learning is not directly applicable. Rather than creating new forms of user interaction, we apply preference feedback~\cite{christiano2017deep,DBLP:conf/emnlp/GaoMG18,ouyang2022training} when faced with multiple candidates. Since there are only two candidates, leaving preference feedback will be as simple as binary feedback. 

For the chosen models $i$ and $j$ at time $t$, the batch of input $\mathcal{B}_t \sim X$ will be given to them independently, to obtain the predicted answer spans. Then the users need to compare the two predictions to indicate a preference, where the more accurate answer should be picked out. 

\noindent{\textbf{Reward.}} \ For each data instance in the batch, the reward $r \in \{0,1\}$. $r=1$ means the user prefers one answer from the two candidates; $r=0$ means the user has no preference -- either the two answers are both good or none is good. This is a strict measurement for preference since the answers without preference are discarded. 


To simulate the preference, we calculate the quality score of the predicted answers against the annotated spans, where the answer with a higher score would be preferred or no preference is made if the scores are the same. We use the index-wise F1 value as the quality score, which calculates the F1 score over the predicted indices and the annotated indices, so the score is continuous from $[0,1]$. 

For the batch of input $\mathcal{B}_t$, the quality score for the model $i$ and $j$ is denoted as a vector $\mathbf{s}_{i}$ and $\mathbf{s}_j$ respectively. The rewards $\mathbf{r}_i$ and $\mathbf{r}_j$ for the model $i$ and $j$ respectively are obtained by:
\begin{equation}
    \mathbf{r}_i = \mathbf{s}_i > \mathbf{s}_j ; \ \  \mathbf{r}_j = \mathbf{s}_j > \mathbf{s}_i
    \label{eq:preference-reward}
\end{equation}
where $\mathbf{r}_i$ and $\mathbf{r}_j$ are one-hot vectors.

\noindent{\textbf{Collaborative Model Update.}} \ After obtaining the rewards, we perform collaborative model updates. If there is one preferred model, then it will be regarded as the teacher, and its prediction will be used to jointly update the two models. With the predictions from model $i$ and $j$, we obtain the better one $\langle \mathbf{y}^{*(1)}_t, \mathbf{y}^{*(2)}_t \rangle $, each as a vector, as:
\begin{align}
    \langle \mathbf{y}^{*(1)}_t, \mathbf{y}^{*(2)}_t \rangle = \langle \mathbf{r}_i \mathbf{y}_t^{(i)(1)} + \mathbf{r}_j \mathbf{y}_t^{(j)(1)}, \notag \ \ \ \ \ \ \ \ \ \  \\ \ \ \ \ \   \mathbf{r}_i \mathbf{y}_t^{(i)(2)} + \mathbf{r}_j \mathbf{y}_t^{(j)(2)} \rangle
\end{align}
where $\mathbf{y}^{*(1)}_t$~($\mathbf{y}^{*(2)}_t$) is a vector of the predicted start~(end) positions from the preferred model for the batch of input $\mathcal{B}_t$. 

Then we jointly update the two models by the loss:
\begin{equation}
    \mathcal{L}_t = (\mathbf{r}_i + \mathbf{r}_j)^\top \mathcal{L}(\mathbf{y}^{*(1)}_t, \mathbf{y}^{*(2)}_t) / |\mathcal{B}_t|
    \label{eq:collaborative-update}
\end{equation}
Models updated in this way can better make use of the benefits of different source models during training, that is, when one model from a selected pair cannot predict a correct answer, the other one may make up for it by sharing its prediction. 

\noindent{\textbf{Inference.}} \ After adaptation, there is no need to access a pair of models for inference anymore, so we just keep the best performing model by the method in line $12$ of Algorithm~\ref{algo:co-ucb}. 

\begin{figure}
\setlength{\abovecaptionskip}{-0.2cm}
\setlength{\belowcaptionskip}{-0.3cm}
    \centering
    \includegraphics[width=5cm]{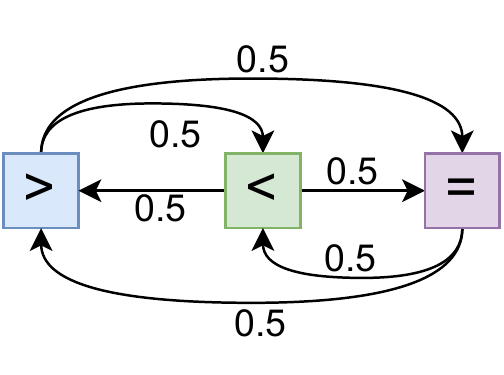}
    \caption{Transition probability for noise simulation, e.g., the correct option `$>$' is corrupted into `$<$' or `$=$' equally by the same probability.}
    \label{fig:noise-transition}
\end{figure}

\subsection{Noise Simulation}
Implicit preference feedback is naturally noisy since the preferred answer only needs to be better than the other and is not necessarily fully correct. However, the users may wrongly provide a preference in practice. Thus, we provide a pilot study to investigate the effect of such noise on adaptation performance. 

There are three options that the user may provide over two candidates, which are `$>$', `$<$'~(with preference), and `$=$'~(no preference). For each data instance, we have a \emph{noise rate} to randomly decide whether its feedback should be corrupted or not. If the feedback should be corrupted, then the correct option is changed to one of the remaining two options with a probability. In this work, we use the transition probabilities shown in Figure~\ref{fig:noise-transition}. We leave more complex transition probabilities to future work.

\begin{figure*}[t]
    \centering
    \includegraphics[width=\textwidth]{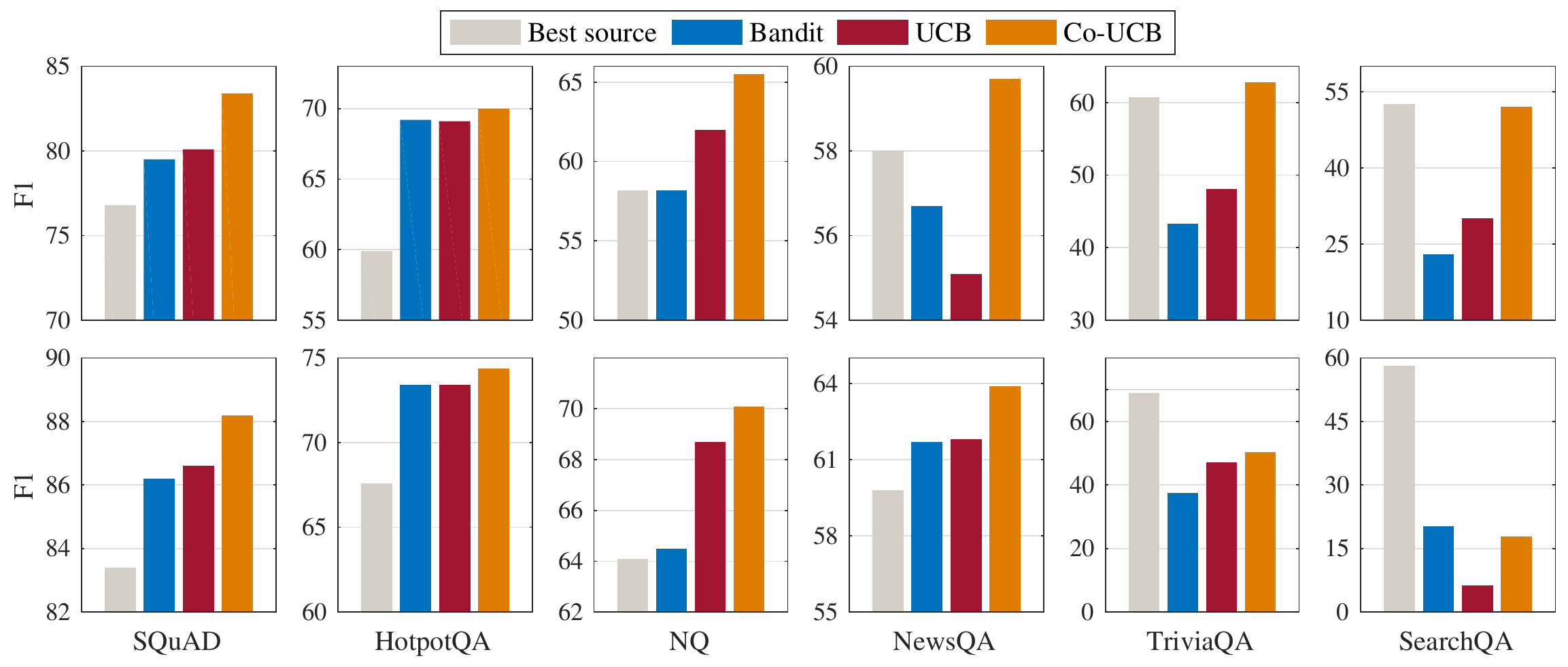}
   \caption{Results of 5-source test-time adaptation. Experiments for the adaptation methods are run three times with random seeds, and the average results are reported. The results of the first row are based on XLMR, and those of the second row are based on SpanBERT.}
\label{tab:main-result}
\end{figure*}

\begin{figure*}[t]
    \centering
    \includegraphics[width=14cm]{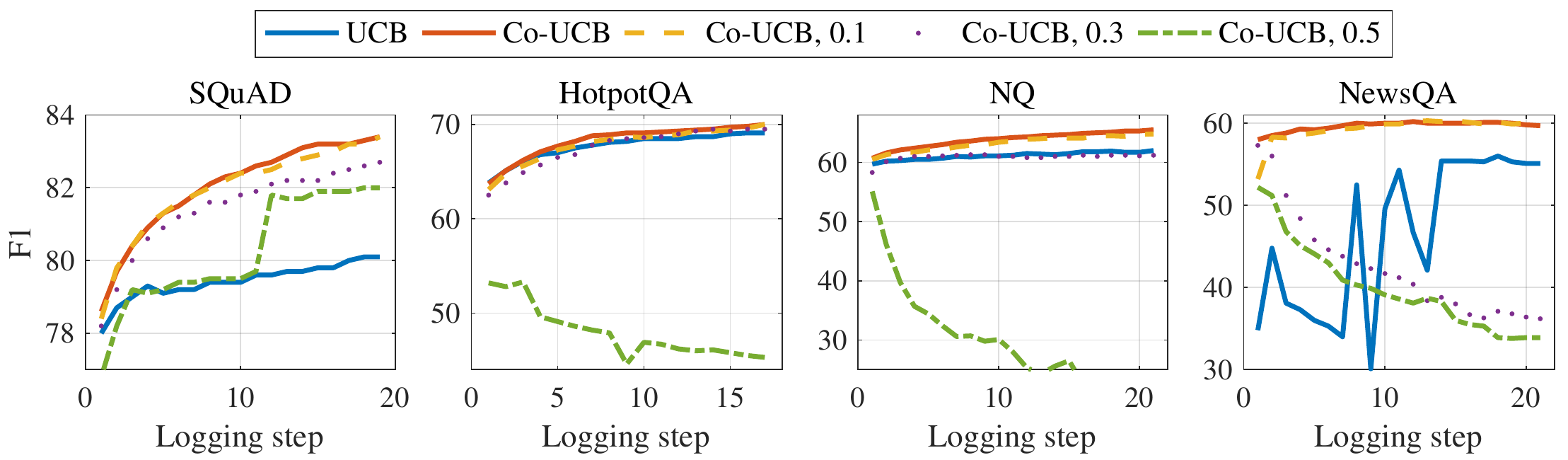}
    \caption{F1 performance on XLMR w.r.t. logging steps. 0.1, 0.3, and 0.5 are the noise rates. Results of TriviaQA and SearchQA are shown in Figure~\ref{fig:logging-rest}.}
    \label{fig:main-draw-noise}
\end{figure*}

\section{Experiments}

\subsection{Simulation Setup}
\noindent{\textbf{Dataset.}} We conduct our experiments on MRQA \cite{fisch2019mrqa}, which is a standard benchmark for domain generalization in extractive QA. We study six datasets~(domains), which are SQuAD~\cite{rajpurkar2016squad}, HotpotQA~\cite{DBLP:conf/emnlp/Yang0ZBCSM18}, Natural Questions~(NQ)~\cite{47761}, NewsQA~\cite{DBLP:conf/rep4nlp/TrischlerWYHSBS17}, TriviaQA~\cite{JoshiTriviaQA2017}, and SearchQA~\cite{DBLP:journals/corr/DunnSHGCC17}, where each dataset forms a distinct domain. The training and development sets are used in our study.

\noindent{\textbf{Setting of 5-source TTA.}} \ To establish multi-source domain adaptation from the six domains, we set each dataset as the target domain and the remaining five datasets as the source domains. For each adaptation, the training set of the target domain is used as the unlabeled test data by discarding the labels, and the development set of the target domain is held out to evaluate the adaptation performance. 

\noindent{\textbf{Evaluation Metric.}} \ We use F1 score to evaluate the performance on the held-out development set. 

\noindent{\textbf{Training Details.}} \ We use the training set of each domain to train each source model, which follows the training details of \citet{hu2020xtreme}. We utilize XLMR~\cite{conneau2019unsupervised} and SpanBERT~\cite{joshi2020spanbert} as the pre-trained language model. In each multi-source domain adaptation, we set the batch size as $16$ and use a constant learning rate of 5e-7. The number of unlabeled test data instances is limited to 100K. The embedding layer is frozen to save computation. Experiments were conducted on one NVIDIA A100 GPU.

\noindent{\textbf{Baselines.}} \ We first present the results of the best source model without adaptation~(\textbf{Best source}). Since our work is the first to study multi-source TTA, there are no existing baselines that address the same problem, so for the multi-source scenario, we mainly compare the two frameworks discussed above. \textbf{UCB} addresses the problem as bandit learning from binary feedback. \textbf{Co-UCB} is for dueling bandits from simulated preference feedback\footnote{Due to the limitation of computing resources, we are not able to train an ensemble model of 5 source models, so we do not show the results of the ensemble model.}. 

We further compare to single-source TTA which has been studied in \citet{gao2022simulating}. We first find the best source model before adaptation by evaluating each model on the held-out development set, then adapt the best source model from simulated binary feedback following the method of \citet{gao2022simulating}. This baseline is denoted as \textbf{Bandit}.

\subsection{Main Results}
We first show the results of 5-source TTA in Figure~\ref{tab:main-result}. First, consistent with the findings of \citet{gao2022simulating}, successful adaption is hard to see on TriviaQA and SearchQA just as the baseline of Bandit~\cite{gao2022simulating} indicates, so the following observations are based on the results of the remaining four target domains.

\noindent{\textbf{Bandit and dueling bandits learning are effective in determining useful sources.}} \ We find both UCB and Co-UCB can effectively improve the adaptation results compared to the best source without adaptation, which indicates that useful sources are found for adaptation during the training process. 

\noindent{\textbf{Leveraging multiple sources by Co-UCB performs the best.}}  
Even without learning a $K$-ensemble model, Co-UCB still improves over the baselines by a large margin. Co-UCB can effectively utilize the benefits of different sources to outperform the Bandit baseline that adapts only one source model. On the contrary, UCB is not effective in making use of multiple sources, since it only achieves results similar to Bandit. 


\noindent{\textbf{Results during adaptation.}} \ Figure~\ref{fig:main-draw-noise} and Figure~\ref{fig:logging-rest} plot the F1 scores vs.~logging steps, where we find that UCB shows a large variance during adaptation on NewsQA, TriviaQA, and SearchQA, i.e., it is slow to find the best source model on these target domains. Co-UCB exhibits better performance and lower variance than UCB during adaptation.

\subsection{Noise Simulation}
\begin{figure}
    \centering
    \includegraphics[width=\columnwidth]{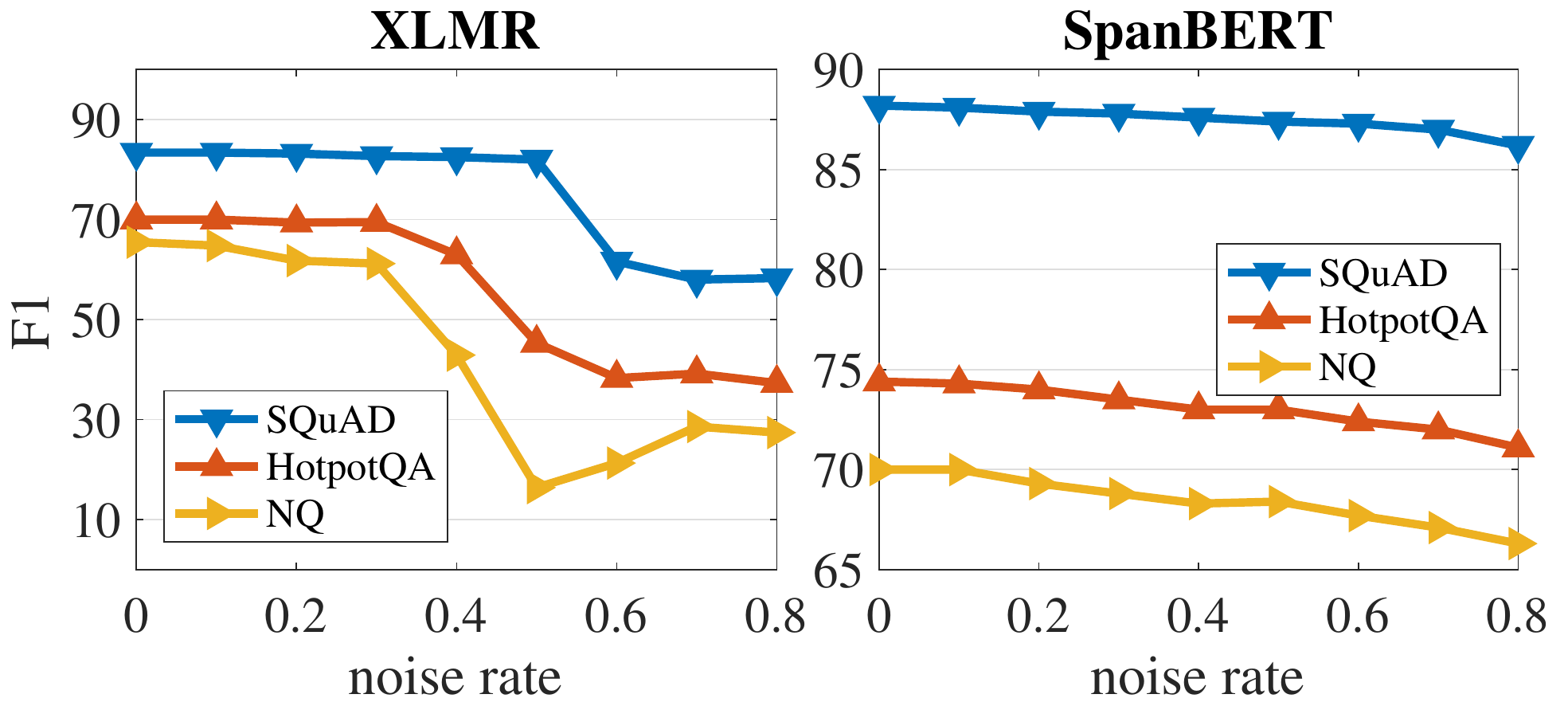}
    \caption{Adaptation results of Co-UCB w.r.t. noise rates using the noise transition in Figure~\ref{fig:noise-transition}.}
    \label{fig:noise-simulation}
\end{figure}

We use the transition probabilities in Figure~\ref{fig:noise-transition} to simulate the noise, e.g., flip the feedback from ``$>$'' to ``$=$'' with probability of $0.5$. 
Results are presented in Figure~\ref{fig:noise-simulation}. As the noise rate increases and more test data's feedback is corrupted, the performance decreases. The result on XLMR drops dramatically with a noise rate larger than $0.5$, but on SpanBERT, the rate of decline is not as fast. SpanBERT is a stronger LM than XLMR for extractive QA, so it is more robust to noisy feedback. As shown in Figure~\ref{fig:main-draw-noise}, a large noise rate~(e.g., 0.5) will make the adaptation fail quickly on XLMR.

\begin{table}[]
    \centering
    \resizebox{\columnwidth}{!}{%
    \Huge
\begin{tabular}{l|l|cccc}
\toprule[2.5pt]
                          & \textbf{Targets}             & \textbf{SQuAD} & \textbf{HotpotQA} & \textbf{NQ}  & \textbf{NewsQA} \\ 
                          & \textbf{Baselines}           & F1             & F1                & F1           & F1              \\ \midrule
\multirow{4}{*}{XLMR}     & UCB                          & 80.1$_{0.1}$   & 69.1$_{0.1}$      & 62.0$_{0.9}$ & 55.1$_{0.7}$    \\ 
                          & Co-UCB                    & 83.4$_{0.1}$   & 70.0$_{0.1}$      & 65.5$_{0.1}$ & 59.7$_{0.1}$    \\ 
                          & \multicolumn{1}{r|}{w/o co.} & 79.3$_{0.1}$   & 69.1$_{0.4}$      & 61.4$_{0.2}$ & 51.9$_{0.4}$    \\ 
                          \midrule \midrule
\multirow{4}{*}{SpanBert} & UCB                          & 86.6$_{0.1}$   & 73.4$_{0.1}$      & 68.7$_{0.3}$ & 61.8$_{0.3}$    \\ 
                          
                          & Co-UCB                    & 88.2$_{0.0}$   & 74.4$_{0.1}$      & 70.1$_{0.3}$ & 63.9$_{0.2}$    \\ 
                          & \multicolumn{1}{r|}{w/o co.} & 85.2$_{0.1}$   & 73.0$_{0.1}$      & 68.2$_{0.2}$ & 61.1$_{0.4}$  \\
                        \bottomrule[2.5pt]
\end{tabular}%
}
    \caption{Results of the ablation study. `w/o co.' means removing the collaborative update in Co-UCB.}
    \label{tab:ablation}
\end{table}

\subsection{Further Analysis}
\noindent{\textbf{Ablation study.}} \ Firstly, as Table~\ref{tab:ablation} shows, without collaborative update, Co-UCB clearly degrades and it cannot compete with UCB. Considering that preference feedback is naturally noisy, Co-UCB without collaboration does not perform better than UCB. 


\begin{table}[]
\setlength{\abovecaptionskip}{0.2cm}
\setlength{\belowcaptionskip}{-0.1cm}
\centering
\Huge
\resizebox{\columnwidth}{!}{%
\begin{tabular}{lcccccc}
\toprule[2.5pt]
\textbf{} & \textbf{SQuAD} & \textbf{HotpotQA} & \textbf{NQ} & \textbf{NewsQA} & \textbf{TriviaQA}  &\textbf{SearchQA} \\ \midrule[2pt]
UCB       & 4.77           & 3.99              & 3.55        & 2.36            & 2.03    &           2.16              \\ 
Co-UCB & \textbf{7.25}           & \textbf{5.53}              & \textbf{8.60}        & \textbf{8.91}            & \textbf{8.11}              & \textbf{8.17}              \\ \bottomrule[2.5pt]
\end{tabular}%
}
\caption{Overall rewards~($\times 10^4$) obtained during adaptation based on SpanBERT.}
\label{tab:overall-reward}
\end{table}

\noindent{\textbf{Overall rewards.}} \ We calculate the overall rewards that UCB and Co-UCB obtain during adaptation in Table~\ref{tab:overall-reward}. The overall rewards are the sum of the cumulated rewards from each model. 
We observe that Co-UCB has a higher reward than UCB. In Co-UCB, for a certain input, when one model could not obtain the reward $1$, the other model may make up for it by sharing, so that this model can also be updated and improved. The results support why Co-UCB performs better than UCB: higher rewards can better instruct the model to adapt. 

\begin{figure}[t]
    \centering
    \includegraphics[width=7cm]{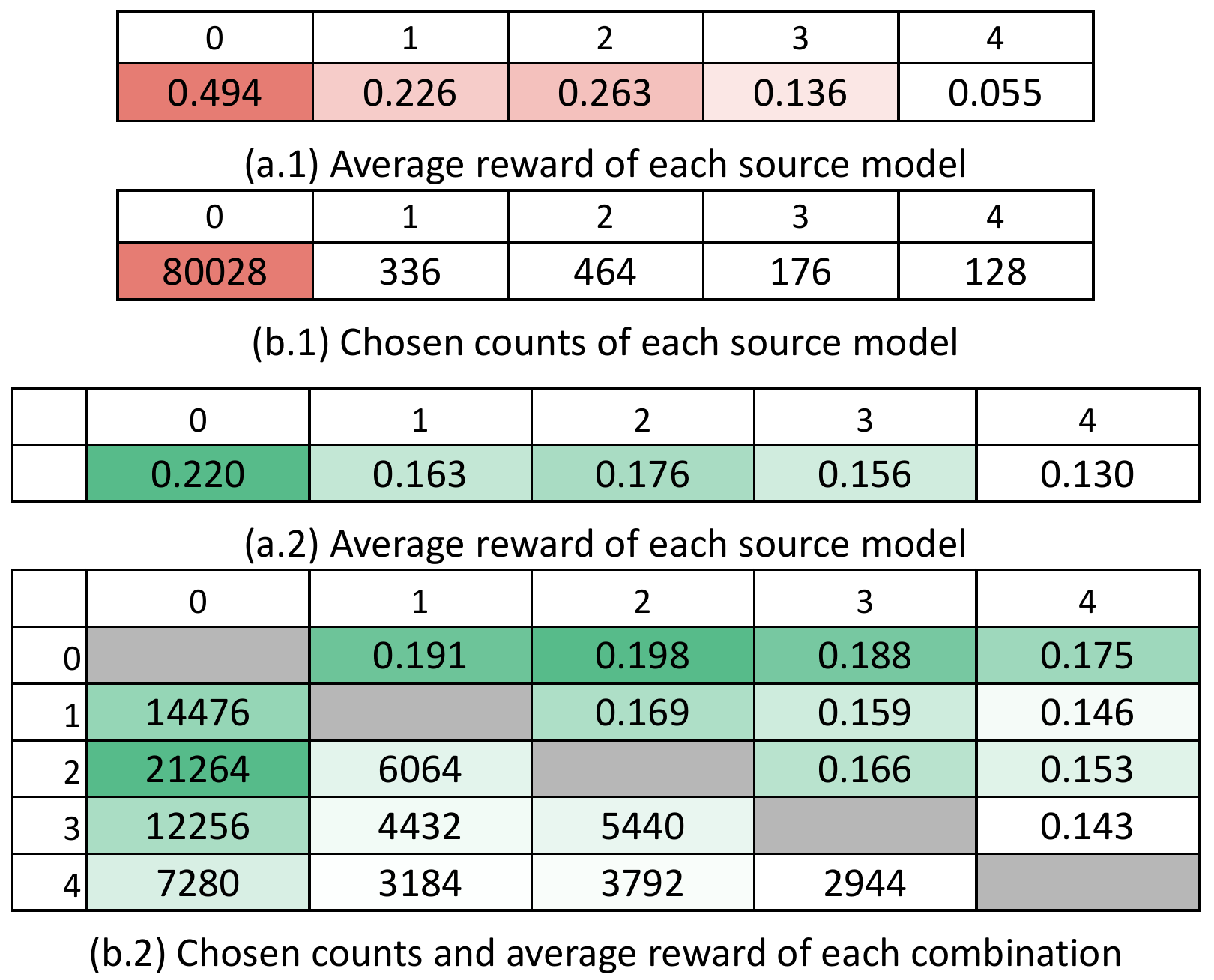}
    \caption{Average rewards and chosen counts of UCB~(a.1, b.1) and Co-UCB~(a.2, b.2) obtained during adaptation on the target of HotpotQA.}
    \label{fig:case-study-1}
\end{figure}

\begin{figure}[t]
    \centering
    \includegraphics[width=\columnwidth]{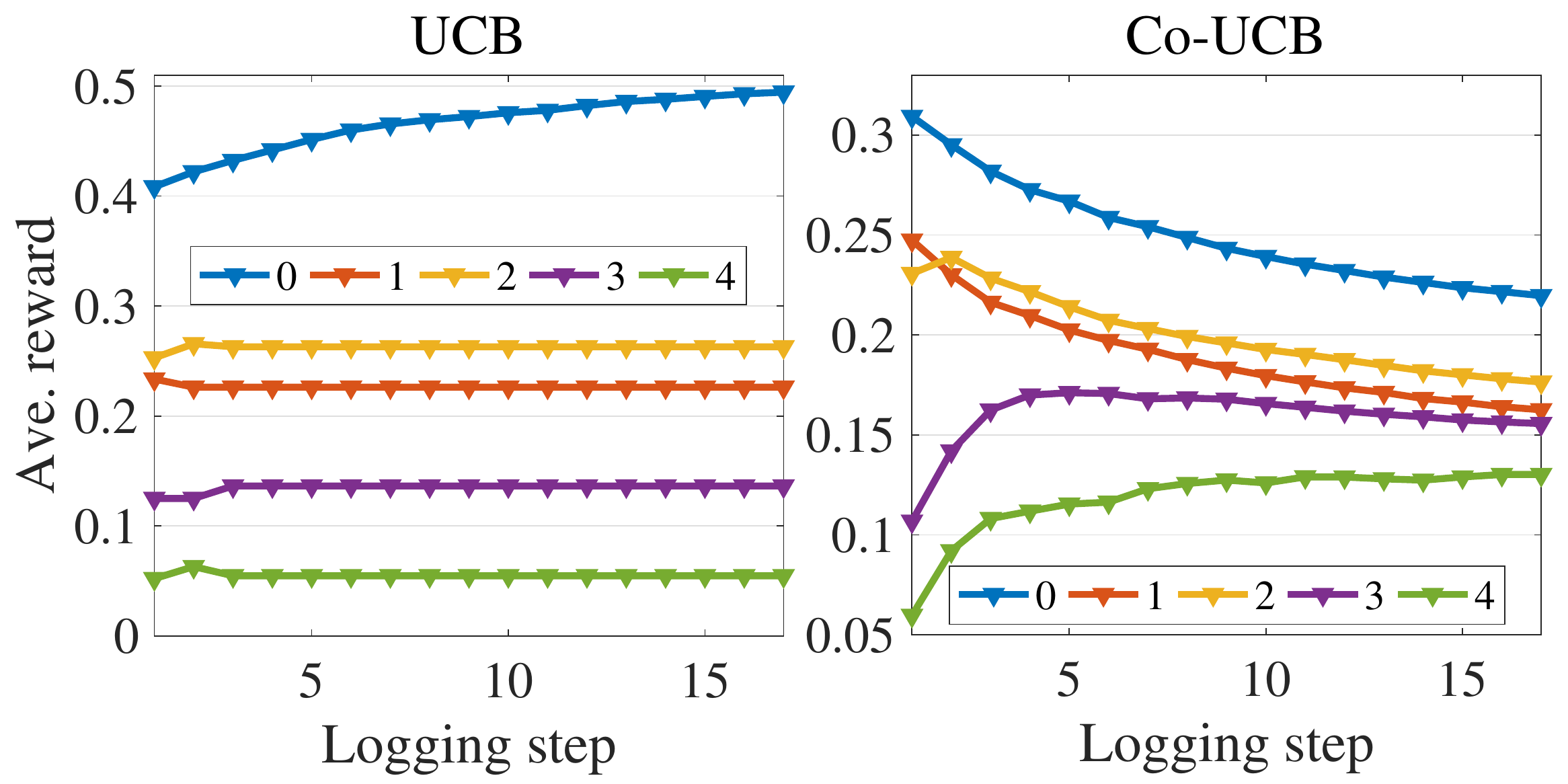}
    \caption{Average reward w.r.t. logging step for each model during adaptation on HotpotQA.}
    \label{fig:case-study-2}
\end{figure}

\noindent{\textbf{Case study.}} \ We show the average reward and chosen count for each source model during adaptation in Figure~\ref{fig:case-study-1}. For Co-UCB, the models $0$ and $2$ are the best and second best model respectively, so its combination $\langle 0, 2 \rangle$ is chosen most of the time. $\langle 0, 1 \rangle $ and $\langle 0, 3 \rangle$ are also often accessed since their payoff is close to the best one. However, UCB would mostly only focus on updating the best model which is model $0$. As shown in Figure~\ref{fig:case-study-2}, UCB is able to quickly find the best source model~(model $0$), and the other models would be discarded without updating. For Co-UCB, since the models are dueling with each other, the changing of rewards behaves differently. The reward of model $0$, $1$, and $2$ decreases, while the reward of model $3$ and $4$ increases, since dueling bandits learning is a zero-sum game in general~\cite{DBLP:conf/colt/YueBKJ09}, where one model winning in dueling means another model loses. However, reward sharing happens in Co-UCB during training.  

\begin{figure}[t]
    \centering
    \includegraphics[width=7cm]{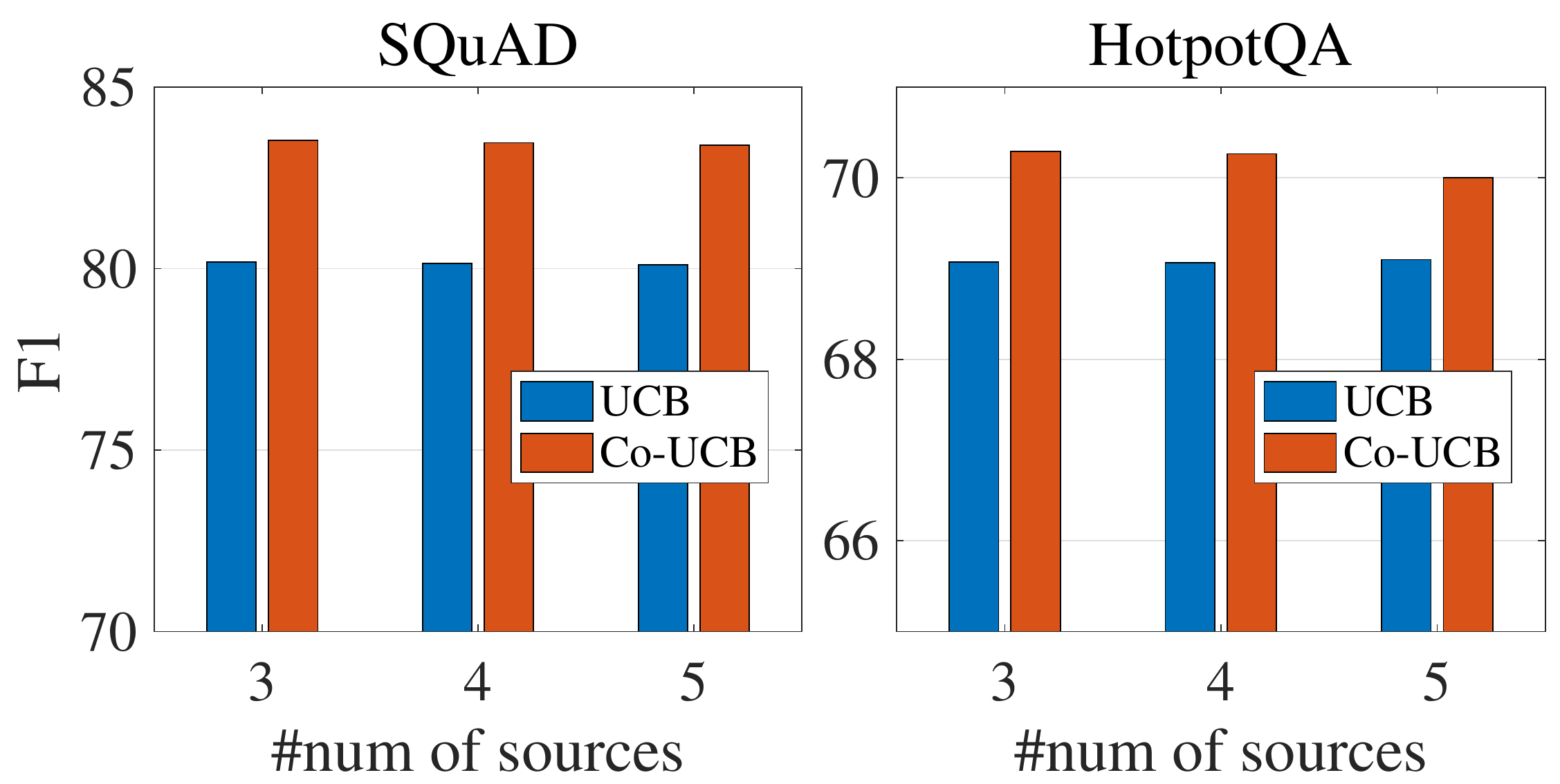}
    \caption{Effects of the number of source models on adaptation performance based on XLMR.}
    \label{fig:num-of-source}
\end{figure}

\noindent{\textbf{Effects of the number of source domains.}} \ 
From the results in Figure~\ref{fig:num-of-source}, we can see that the adaptation results have a very slight drop when the number of sources increases. No matter how the number of source models changes, Co-UCB still consistently performs better than UCB. 

We also discuss the effects of preference feedback on UCB in Appendix~\ref{sec:ucb_preference}.

\section{Conclusion}
We present the first work on multi-source test-time adaptation from human feedback, where we cast it as an iterative decision-making problem. We first formulate it as multi-armed bandit learning. More importantly, to utilize pairwise collaboration, we further regard it as dueling bandits. Co-UCB is a novel method proposed in this work. Though we study online adaptation from the online data stream, our work can also be applied in offline model refinement. For the offline setting, we do not update the model online but only update the policy for model selection when receiving user feedback each time. After collecting enough user feedback, we fine-tune the found best model offline with user feedback.   

\section*{Limitations}
Learning an ensemble of multiple source models is expensive, especially for large language models. Hence, to adapt to the new target domain, we cast the problem as an iterative-decision making process. While our work reduces the model access frequency to $1$ or $2$ at each training step, continually updating the language model from a stream of test data is still costly. Future work can explore better methods for efficient optimization for a single LM. Besides, in some cases, the distribution of test data may change dynamically over the stream, but our work considers only the situation where the test data is from one specific distribution. More complex cases of test distribution can be studied in future work.

\section*{Acknowledgements}
This research is supported by the National Research Foundation, Singapore under its AI Singapore Programme (AISG Award No: AISG-RP-2018-007 and AISG2-PhD-2021-08-016[T]). The computational work for this article was partially
performed on resources of the National Supercomputing Centre, Singapore (https://www.nscc.sg).

\bibliography{anthology,custom}
\bibliographystyle{acl_natbib}


\appendix

\begin{figure*}
    \centering
    \includegraphics[width=12cm]{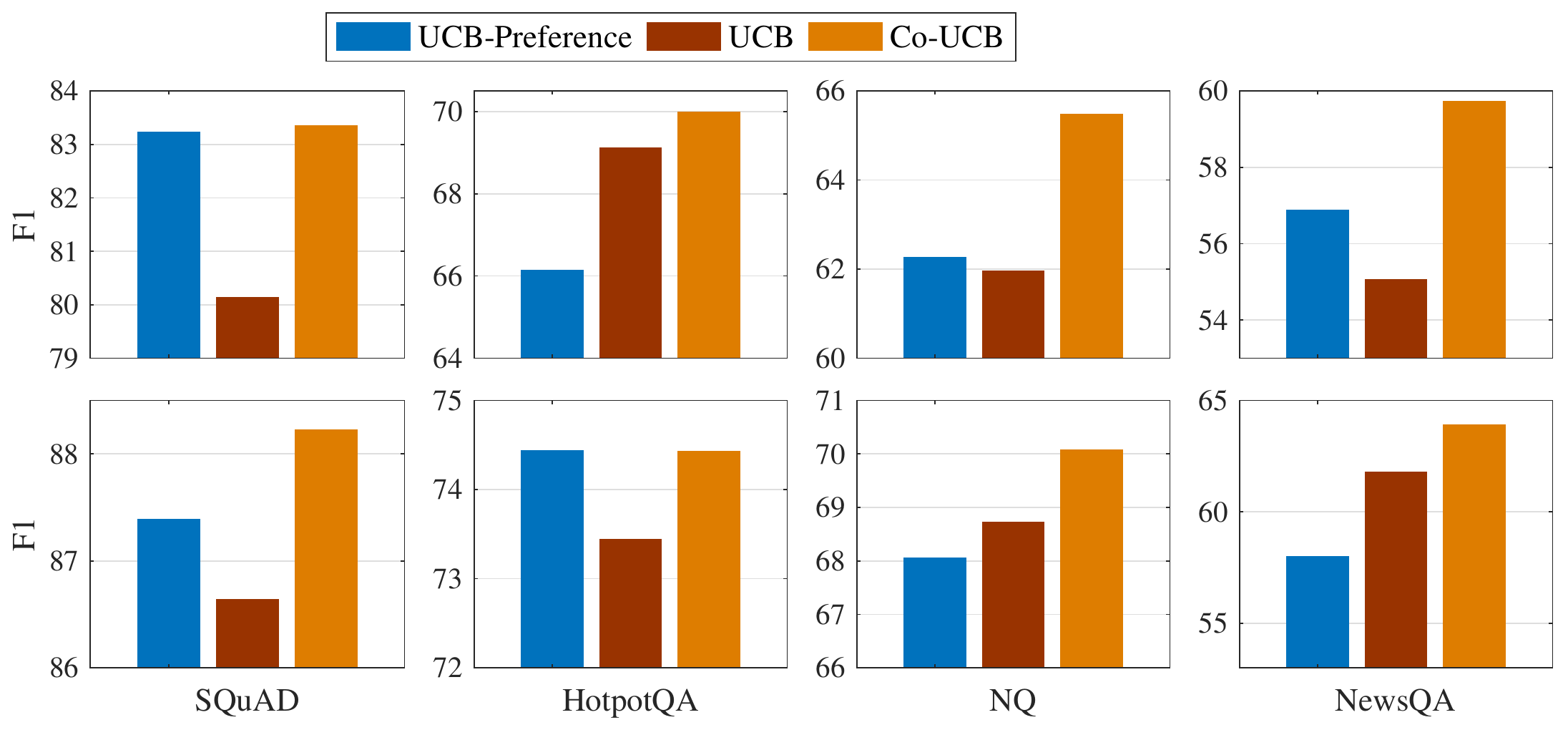}
    \caption{The results of UCB with preference feedback~(i.e., UCB-Preference) with comparison to UCB and Co-UCB. The first row is based on XLMR, and the second row is based on SpanBERT. Experiments are run three times with random seeds, and the average results are reported.}
    \label{fig:ucb_preference}
\end{figure*}

\section{Effects of Preference Feedback on UCB}\label{sec:ucb_preference}
In the main content of this paper, we use preference feedback for Co-UCB, since each time the user has a pair of predictions to provide feedback. For UCB, the user only has one candidate to leave feedback, so we use binary feedback.

Here, we further study how preference feedback would affect the performance of UCB. To enable preference feedback, for each input data instance, the model first generates its top two predictions~(to be comparable to Co-UCB), then the user needs to provide preference feedback to the two candidates. We follow the same procedure of Co-UCB to simulate preference feedback for UCB. 

The results are presented in Figure~\ref{fig:ucb_preference}. As we can see, UCB with preference feedback improves over UCB with binary feedback in some cases~(not a consistent improvement), since top two predictions give the user more choices to select a good label. However, UCB with preference feedback cannot compete with Co-UCB. Co-UCB aims to leverage the benefits of different source models instead of the model's top several predictions, which is different from UCB with preference feedback. Similar to UCB with binary feedback, UCB with preference also lacks collaboration among source models, since the top two predictions, though expanding the options to select a good label, are just from one model. This finding further demonstrates the effectiveness and importance of leveraging multiple source models during test-time adaptation. 


\begin{figure}
\setlength{\abovecaptionskip}{0.1cm}
\setlength{\belowcaptionskip}{-0.1cm}
    \centering
    \includegraphics[width=\columnwidth]{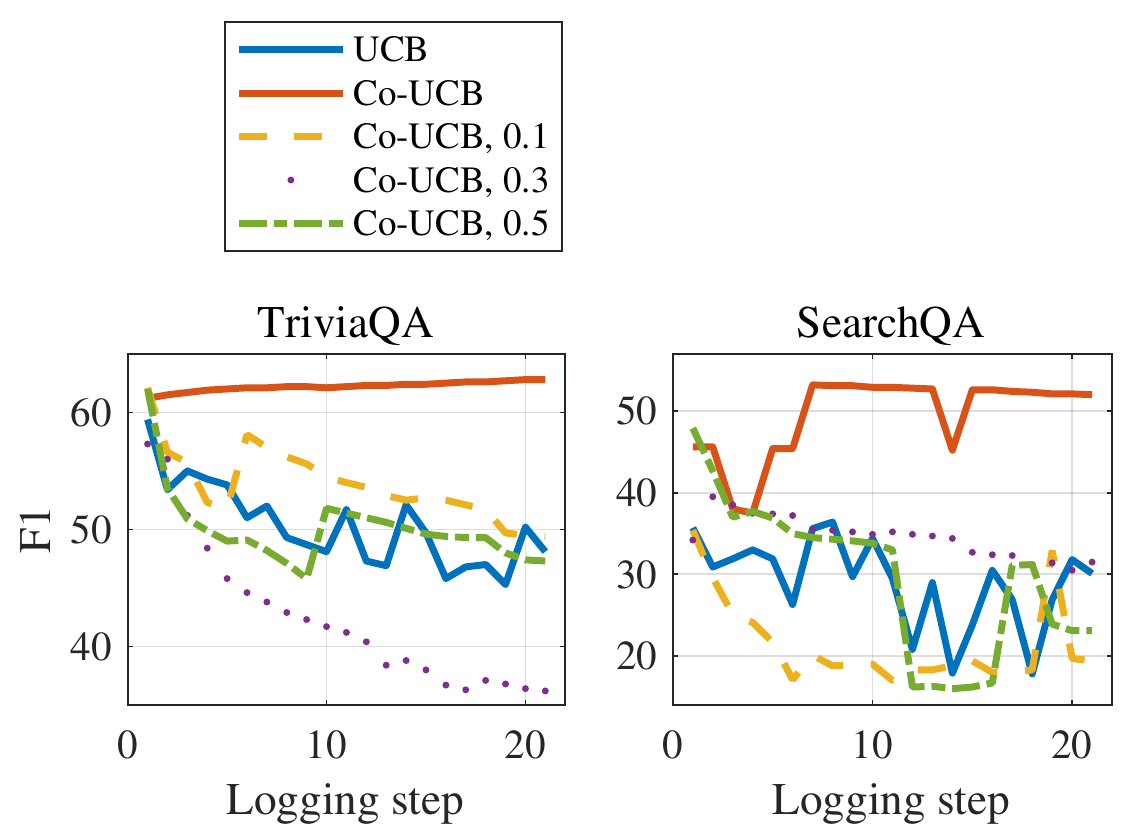}
    \caption{F1 performance on XLMR w.r.t. logging steps. 0.1, 0.3, and 0.5 are the noise rates.}
    \label{fig:logging-rest}
\end{figure}

\end{document}